\documentclass[10pt,twocolumn,letterpaper]{article}

\usepackage{iccv}
\usepackage{times}
\usepackage{epsfig}
\usepackage{graphicx}
\usepackage{amsmath}
\usepackage{amssymb}

\usepackage{subfiles}
\usepackage[group-separator={,},output-decimal-marker = {.}, group-minimum-digits={3}]{siunitx}  %
\usepackage{paralist} %
\usepackage{booktabs} %
\usepackage{multirow}
\usepackage{algpseudocode}
\usepackage{algorithm}
\usepackage{tikz}
\usetikzlibrary{mindmap,trees}
\usepackage{verbatim}
\usepackage{pgfplots}
\pgfplotsset{compat=newest}
\usepackage{csvsimple}
\usepackage{braket}

\usepackage[pagebackref=true,breaklinks=true,letterpaper=true,colorlinks,bookmarks=false]{hyperref}

\newcommand{\putpin}[4]{\addplot[forget plot,mark=none] coordinates {#1} node[pin={[pin distance=#2,inner sep=0pt,outer sep=1pt,fill=white,label={[inner sep=0pt]#4}]#3:{}}]{};}
\newcolumntype{H}{>{\setbox0=\hbox\bgroup}c<{\egroup}@{}}

\newcommand{\vitto}[1]{}

\newcommand{\att}[1]{#1}

\newcommand{\coco}{COCO}
\newcommand{\cocostuff}{COCO-Stuff}

\newcommand{\openimages}{Open~Images}

\renewcommand{\paragraph}[1]{\vspace{4pt}\noindent\textbf{#1}}

\iccvfinalcopy %

\ificcvfinal\fi
\begin{document}

   \title{\vspace{-8mm}Natural Vocabulary Emerges from Free-Form Annotations\vspace{-4mm}}

   \author{Jordi Pont-Tuset\hspace{5mm}Michael Gygli\hspace{5mm}Vittorio Ferrari\\{\tt\small \{jponttuset, gyglim, vittoferrari\}@google.com}\\[0.9mm]{Google Research}
}

\maketitle

\begin{abstract}
   We propose an approach for annotating object classes using free-form text written by undirected and untrained annotators.
   Free-form labeling is natural for annotators, they intuitively provide very specific and exhaustive labels, and no training stage is necessary.
   We first collect \att{\num{729}k} labels on \att{\num{15}k} images using \att{\num{124}} different annotators.
   Then we automatically enrich the structure of these free-form annotations by discovering a \textit{natural vocabulary} of \att{\num{4020}} classes within them.
   This vocabulary represents the natural distribution of objects well and is learned directly from data, instead of being an educated guess done before collecting any labels.
   Hence, the natural vocabulary {\em emerges} from a large mass of free-form annotations.
   To do so, we (i) map the raw input strings to entities in an ontology of physical objects (which gives them an unambiguous meaning);
   and (ii) leverage inter-annotator co-occurrences, as well as biases and knowledge specific to individual annotators.
   Finally, we also automatically extract natural vocabularies of reduced size that have high object coverage while remaining specific.
   These reduced vocabularies represent the natural distribution of objects much better than commonly used predefined vocabularies.
   Moreover, they feature more uniform sample distribution over classes.
\end{abstract}

  \vspace{-6pt}
  \section{Introduction}\label{sec:introduction}
  When tackling the problem of localizing multiple objects in images, the first question one faces is \textit{which classes of objects?}
  Annotated datasets in this field~\cite{everingham15ijcv,lin14eccv,russakovsky15ijcv,kuznetsova18arxiv,openimages} ask annotators to mark any instance of a predefined set of classes (\eg{} person, dog, etc.), which we refer to as \textit{predefined vocabulary}.
  The choice of such vocabulary has a large impact on the algorithms developed on these datasets.
  Despite this, it has always been an arbitrary and predefined choice.
  
   \begin{figure*}
    \centering
    \resizebox{0.9\linewidth}{!}{%
    \includegraphics{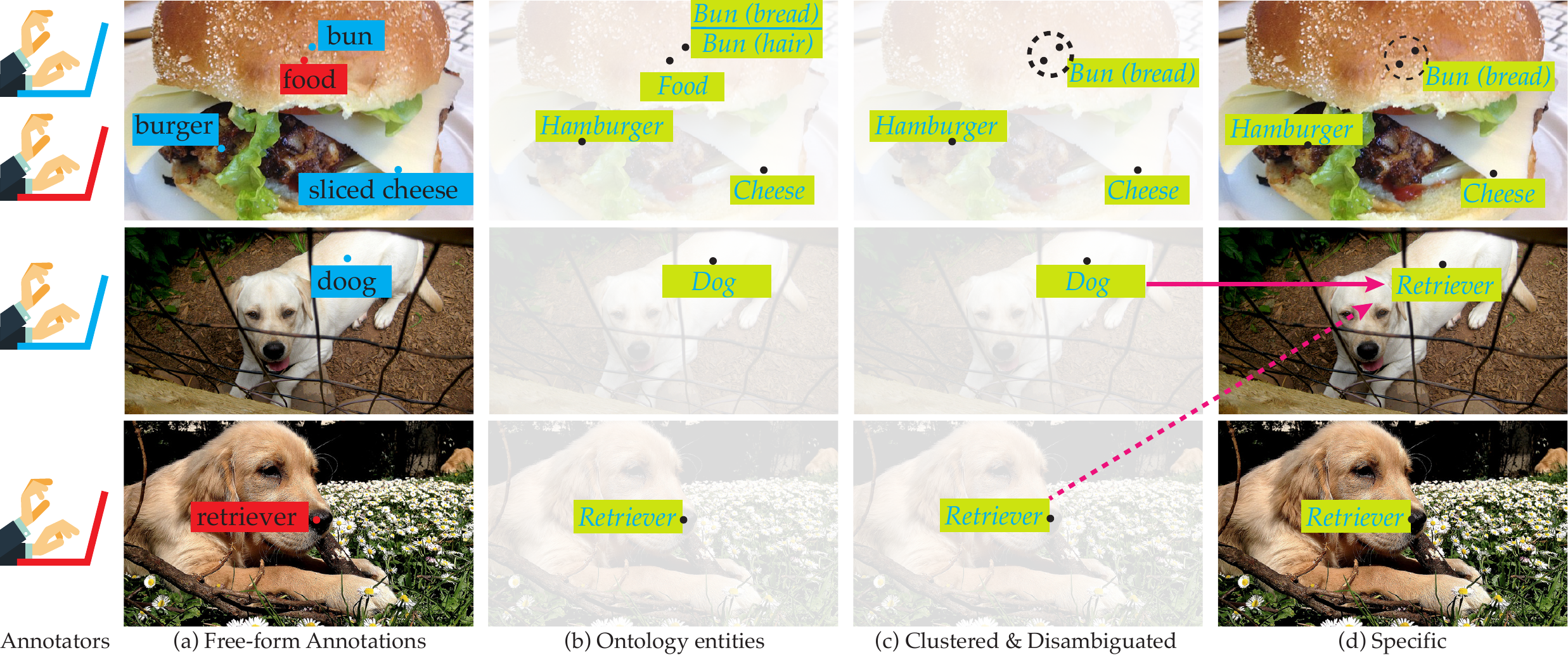}}
    \caption{\textbf{Method overview}: Various annotators (red and blue) annotate points and label them using free-form text (a).
    Our method then normalizes the strings into entities of a visual hierarchy (b), with potentially ambiguous results.
    We spatially cluster the points on each image to detect groups of points belonging to the same object (c), which in turn is used to disambiguate meanings (c).
    We finally make labels more specific by combining cross-annotator knowledge (d).\vspace{-8pt}}
    \label{fig:overview}
  \end{figure*}

  We instead propose letting the annotators click on the objects present in the image and label them using free-from text: no predefined
  vocabulary and no policies about what to label (Fig.~\ref{fig:overview}).
  The main contribution of this paper is a principled automatic approach to make a \textit{natural vocabulary} emerge from these points with free-form annotations, \ie{} the vocabulary that best represents all the labeled objects.
  The natural vocabulary is thus learnt from data rather than predefined before any annotation is collected.

  Free-form annotations come with their own challenges, however: spelling mistakes (\textit{doog} in Fig.~\ref{fig:overview}), ambiguous meanings of words (\textit{bread bun} \vs \textit{hair bun} in Fig.~\ref{fig:overview})
  , annotators using different words for the same concept (\textit{bun} \vs \textit{bread} in Fig.~\ref{fig:overview});
  some knowing about car brands, others about dog breeds (\textit{dog} \vs \textit{retriever} in Fig.~\ref{fig:overview});
  some being exhaustive and consistent, others sparse and unsystematic.
  Most critically, free-form annotations lack a structure that facilitates learning from them.
  Finding a natural vocabulary adds structure on the annotations and makes annotator-specific biases and knowledge visible (Sec.~\ref{subsec:matching-to-a-language-vocabulary}-\ref{subsec:natural_vocab}).

We leverage the different expertise of different annotators discovered to automatically make some annotations as specific as possible (Sec.~\ref{subsec:specialization-method}).
For example, we can automatically label specific dog breeds on those objects labeled simply as \textit{dog} by taking into account the knowledge of the annotators that know about dog breeds, even though they did not annotate that specific image (\eg retriever in Fig.~\ref{fig:overview}).

        Furthermore, our proposed annotation scheme is natural for annotators: their brain translates the objects they see into words instantaneously.
    They do not need to be
        trained for the task (in contrast to previous approaches~\cite{su12aaai, lin14eccv, russakovsky15ijcv, gygli19cvpr}), which positively affects the availability and cost of the crowd workforce.

  We annotate \att{\num{729469}} points on \att{\num{15100}} images of \openimages{}~\cite{kuznetsova18arxiv} using \att{\num{124}} different annotators (\num{5} to \num{10} per image).
   The key findings of these experiments are:
  \begin{asparaitem}

    \item We can effectively and automatically extract the meaning of raw input strings, by matching them to concepts in an ontology, with an accuracy of \att{\num{89}\%}.

    \item Our method makes a large natural vocabulary of \att{\num{4020}} classes emerge from the annotations, which is well beyond
    the predefined vocabulary of previous datasets with object location annotations~\cite{everingham15ijcv,lin14eccv,kuznetsova18arxiv}.
    Our natural vocabulary covers both classes beyond these predefined vocabularies and classes more specific than those.

    \item The annotations are exhaustive: they cover \att{\num{7.5}} different classes per image in
    mean using a single annotator per image, and up to \att{\num{19.7}} when using \att{\num{5}}.

    \item We extract reduced vocabularies that represent the natural distribution of objects significantly better than previous predefined vocabularies.
    This allows to cover more objects naturally present in images with a given vocabulary size.

    \item Finally, we also show that the natural vocabulary can be used to train textual-visual embedding models. These offer better zero-shot classification performance, compared to models trained on a fixed vocabulary of \num{600} classes (Sec.~\ref{sec:learning-from-open-vocabulary-annotations}).

  \end{asparaitem}

  \section{Related Work}
  \paragraph{Datasets:}
  The vast majority of datasets label objects from a predefined vocabulary~\cite{fei-fei:pami06,
  krizhevsky09, everingham15ijcv, russakovsky15ijcv, lin14eccv, bossard14eccv, kuznetsova18arxiv}.
  Annotators thus typically need to be trained to provide precise labels within this vocabulary~\cite{su12aaai, lin14eccv, russakovsky15ijcv, gygli19cvpr}.

  Notable exceptions not relying on a predefined vocabulary are ADE20k~\cite{zhou19ijcv} and LabelMe~\cite{russell08ijcv}.
  ADE20k is labeled with a vocabulary that is created on demand, as the annotator encounters objects of previously unseen classes.
  By employing a single expert annotator and a (expandable) vocabulary, inconsistent naming can be minimized, but this approach does not scale to annotating by the crowd.
  Instead, LabelMe is annotated with free-form text labels by multiple crowd workers~\cite{russell08ijcv}, but suffers from inconsistent naming as a result.
  To solve this,~\cite{russell08ijcv} manually matches the collected annotations to an ontology, which allows to link different object labels into a single concept, \eg{} \textit{taxi} and \textit{automobile} to the concept \textit{Car}.

  Different from these techniques, our method provides a principled approach to automatically match free-form annotations to concepts in an ontology and build a natural vocabulary that best represents the input annotations.

  \paragraph{Vision and language:}
    Several areas of Computer Vision directly work with free-form text.
  In image captioning and retrieval, for example, most methods are trained from images with accompanying free-form sentences~\cite{fang15cvpr,donahue15cvpr,xu15iclr, hinami17arxiv, plummer18arxiv}.
\cite{donahue15cvpr} and~\cite{xu15iclr} use recurrent neural networks to generate image captions.
\cite{ling17nips} trains an image caption model using natural language feedback as a training signal.
Several works \cite{guadarrama14rss,hinami17arxiv,plummer18arxiv, rupprecht18cvpr} propose models for localizing objects given free-form text as an input.
  Also related is the task of generating referring expressions~\cite{mitchell10inlg,kazemzadeh14emnlp,mao16cvpr}.
These methods take an image and a bounding box as input and generate a natural language expression, which uniquely localizes an object in an image.

While the methods above learn a mapping from text to images or vice-versa, in our work we learn a natural vocabulary of classes from free-form text annotations on images,~\ie we extract structure.

There are also works that exploit natural language captions to learn visual concepts appearing in the accompanying images.
 In~\cite{berg04cvpr,guillaumin08cvpr}, captions of images in news articles are used to extract labels for faces.~\cite{luo09nips} extracts names and actions from captions and associates them to the faces and body poses of people shown in the images, to train face and pose models.
 In a similar spirit,~\cite{ye18arxiv} proposes a method to detect objects from captions.
  While these methods directly use the raw, noisy data, we propose a method to clean and homogenize annotations into a natural vocabulary.
  Moreover, we show how to detect the biases of the annotators and how to automatically propagate their different knowledge across images.

  \paragraph{Leveraging the crowd:}
  Some previous works extract consensual annotations from multiple annotators, going beyond simple majority voting, \eg{} by treating each annotator differently~\cite{zheng17vldb, guan17aaai}, extending it for multiple-label answers~\cite{tam16tr} and to images~\cite{kim18cvpr}, and classifying annotators into communities~\cite{moreno15jmlr}.
All these works, however, assume there is a clear question to be asked.
Our work, in contrast, lets the annotators choose the question they want to answer to maximize the chances of the responses to be correct.

  \paragraph{Using an object ontology:}
  Several works make use of the WordNet ontology.
  Ordonez~\etal{}~\cite{ordonez13iccv} predict a mapping from WordNet synsets (and an image) to the the names people would naturally use for objects (entry-level classes).
  They thus tackle the inverse problem of ours: to go from commonly used names to a structured vocabulary.

  \cite{russell08ijcv, zhao17iccv} also match free-form text to an ontology (WordNet), but do so manually. \cite{zhao17iccv} matches them in order to obtain a hierarchical representation of labels, which they use to learn hierarchical embeddings.
  Instead, \cite{russell08ijcv} uses the ontology to link different annotations into a single concept, as discussed above.

\section{Method}
\label{sec:method}

Figure~\ref{fig:overview} shows the overview of the proposed method:
(a) we collect free-form text annotations (Sec.~\ref{subsec:click-and-type}),
and (b) match them to possible meanings in a language ontology (Sec.~\ref{subsec:matching-to-a-language-vocabulary}).
We then (c) cluster the points and find their meaning (Sec.~\ref{subsec:matching-between-annotators}-\ref{subsec:postproc}).
Our method makes a natural vocabulary emerge from these annotations (Sec.~\ref{subsec:natural_vocab}) and then (d) uses it to make the annotations as specific as possible (Sec.~\ref{subsec:specialization-method}).

\subsection{Click \& Type}\label{subsec:click-and-type}
We ask the annotators to click on each object present in the image and then type the word(s) that best describe it (Fig.~\ref{fig:overview}a).
We encourage annotators
to be as specific as possible, e.g. type \textit{labrador} if they happen to know the breed or type \textit{animal} if it is an animal they do not recognize.
No other guidelines or policies are given.

As we will see in following sections, we will use the point locations to estimate when two annotators refer to the same object (Sec.~\ref{subsec:matching-between-annotators}) and from that disambiguate the meaning of the words and detect rater biases (Sec.~\ref{subsec:sense-disambiguation}).

\subsection{Matching to a language ontology}
\label{subsec:matching-to-a-language-vocabulary}

The string associated with each point from the \textit{Click \& Type} step is in raw form: a string
from which we have not extracted any meaning or sense.
We can assume, however, that all such strings represent words in the English language, so the next steps in our algorithm
fix spelling errors, match to a possible meanings from a language ontology, etc.

\paragraph{Spelling corrections:}
Correcting spelling mistakes modifies the input strings slightly so that as many strings as possible are valid English words (\eg{} \textit{doog}$\rightarrow$\textit{dog}, Fig.~\ref{fig:overview}).
We use an off-the-shelf English spelling correction algorithm. %

\paragraph{Part-of-Speech tagging:}
Since we do not give annotators any strong policy on which form the strings should take, we might have some adjectives or modifiers in the description of the point (\eg{} \textit{sliced cheese}, Fig.~\ref{fig:overview}).
We run a Part-of-Speech (PoS) tagging algorithm to detect the main noun of the string and separate its modifiers.
We will use this information in the next step.

\paragraph{Entity identification:}
Next, we match the main noun to a language ontology: we select a set of candidate meanings (\textit{entities} in the ontology) from that noun (\eg{} \textit{bun} can be a \textit{Hair bun} or a \textit{Bread bun}, Fig.~\ref{fig:overview}).
We then also match the whole string to the ontology to see if there is a more appropriate candidate (\eg{} a \textit{Sea lion} is not a type of \textit{Lion}).

\paragraph{Restriction to a hierarchy of physical objects:}
The set of candidate entities from the ontology can refer to a very broad spectrum of concepts beyond physical objects directly depictable in an image.
We therefore restrict the possible entities to those in a hierarchy of physical objects: a curated set of classes of physical objects organized in a hierarchical structure where the children nodes refer to subclasses of their parent node.

\subsection{Clustering points into objects}
\label{subsec:matching-between-annotators}

Different annotators might use different words for the same object (\eg{} \textit{bun} and \textit{bread} in Fig.~\ref{fig:overview})
and we want to take advantage of that to disambiguate meanings (Sec.~\ref{subsec:sense-disambiguation}).
To do so, we have the same image labeled by more than one annotator (red and blue in Fig.~\ref{fig:overview}) and then we detect when two points from different annotators refer to the same object.

\begin{figure}
\resizebox{\linewidth}{!}{%
\includegraphics{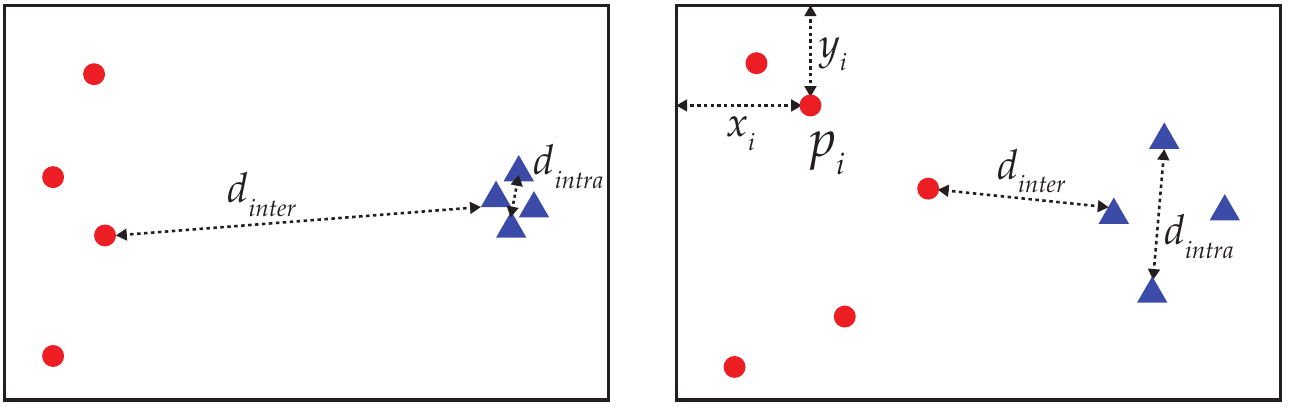}}\\[-2mm]
\mbox{\begin{minipage}{0.5\linewidth}
\centering\footnotesize (a)
\end{minipage}\begin{minipage}{0.5\linewidth}
\centering\footnotesize (b)
\end{minipage}}
\caption{\textbf{Point Clustering}: We look for groups of clicks that are close together and separated from the rest (a) as opposed to more separated points and close to the rest (b).
}
\label{fig:point_clustering}
\end{figure}

Intuitively, we propose to group together points in an image that are close to each other and far from the rest using agglomerative clustering~\cite{rokach05}. We then assume that points in the same cluster refer to the same object (Fig.~\ref{fig:point_clustering}).

As a side effect, we can discover the meaning of words that were not in the language ontology, by finding point clusters where these co-occur with points with unambiguous meaning.

Formally, let $p_i=(x_i, y_i)$ be the $i$th point on an image, regardless of the annotator that provided it.
$x_i, y_i\in[0,1]$ are the normalized (horizontal and vertical) coordinates of~$p_i$ (Fig.~\ref{fig:point_clustering}).
Starting by each point defining its own cluster $c_i=\{p_i\}$, we iteratively merge the two clusters
whose merged result $c$ maximizes the ratio $\mathcal{O}(c)$ defined as:
{\small
\begin{equation}
  \mathcal{O}(c)=\frac{d_{inter}(c)}{d_{intra}(c)}=\frac{\min_{\substack{p_i\in c\\p_j\notin c}}d_E(p_i,p_j)}{\max_{\substack{p_i\in c\\p_k\in c\\}}d_E(p_i,p_k)},
\end{equation}}%
where $d_E(p_i,p_j)=\sqrt{(x_i\!-\!x_j)^2+(y_i\!-\!y_j)^2}$ is the Euclidean distance between $p_i$ and $p_j$.
In theory, $\mathcal{O}(c)$ ranges from $0$ (when there is a point outside $c$ at the same position than a point in $c$) to $+\infty$ (when all points in $c$ are the same and the closest external point is different).

We assume a higher value of $\mathcal{O}(c)$ means a higher probability for all points in $c$ to belong to the same object. Hence, we take all clusters above a certain $\mathcal{O}(c)$ threshold value as same-object points.
In Figure~\ref{fig:point_clustering}a $\mathcal{O}(c)\!\approx\!10$, so we would consider all blue triangles to belong to the same object
(\eg{} \textit{bun} and \textit{bread} in Fig.~\ref{fig:overview}).
Instead, in Figure~\ref{fig:point_clustering}b, $\mathcal{O}(c)\!\approx\!1$ so we discard the blue cluster.

\subsection{Assigning meaning to point clusters}
\label{subsec:sense-disambiguation}

The annotations in each point cluster might come from different annotators and they might use different words, but we assume they all refer to the same object.
Intuitively, we want to assign a meaning to each point cluster that is compatible with all annotations in it.
As an example, the point cluster in Fig.~\ref{fig:overview} contains a \textit{bun} annotation and a \textit{food} annotation.
In isolation, \textit{bun} can refer to the entity \textit{Hair bun} or \textit{Bread bun}, but only \textit{Bread bun} is compatible with \textit{Food}.

To leverage this principle beyond a single cluster, we compute a co-occurrence graph~$G$ where a vertex $v_i$ represents a set of points that are matched to the same set of possible ontology entities (\eg{} one vertex would contain all points matched to \textit{Bread bun or Hair bun}, another node all those matched to \textit{Bread bun} unambiguously, etc.).
We also consider strings that do not match to any entity and add one vertex in the graph per unique string, which we refer to as \textit{unrecognized} vertices.

\begin{figure}
  \centering
  \resizebox{0.9\linewidth}{!}{%
  \includegraphics{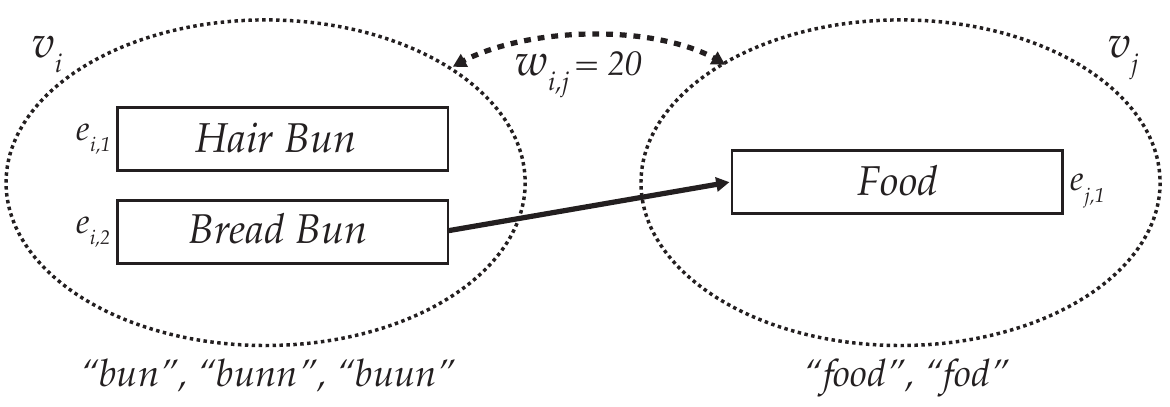}}\\[2mm]
  \resizebox{0.9\linewidth}{!}{%
  \includegraphics{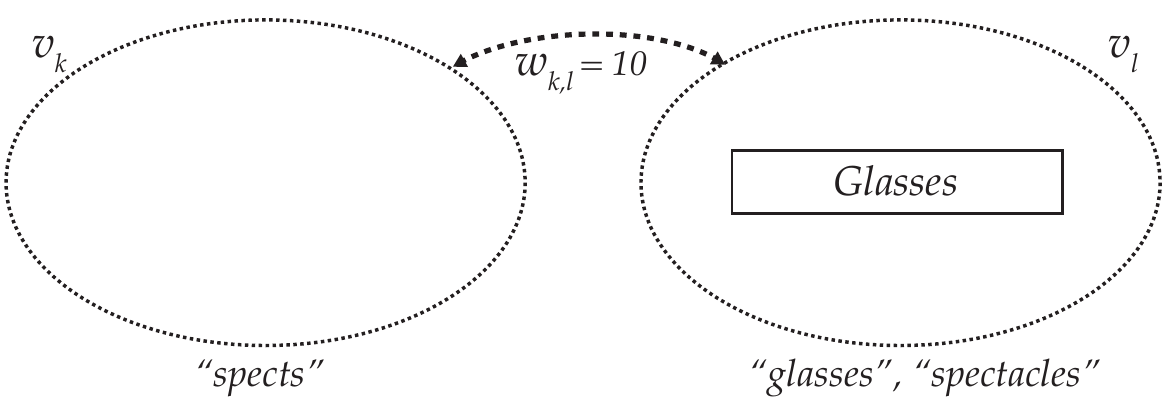}}
  \caption{\textbf{Assigning meaning to clusters}: All points matched to a set of entities $e_{i,x}$ get grouped into a vertex $v_i$.%
  There are $w_{i,j}$ point clusters with at least one point matched to $v_i$ and another to $v_j$.
  The solid arrow connects $e_{i,2}$ to $e_{j,1}$ because the former is a subclass of the latter.
  $v_k$ is is an unrecognized vertex (no meaning found for \textit{spects}).
  \vspace{-8pt}}
  \label{fig:clusters}
\end{figure}

We connect two vertices $v_i$ and $v_j$ with an edge when there are some point clusters with at least one point on each of the connected vertices, and we set its weight $w_{i,j}$ to the number of times this happens in the dataset (Fig.~\ref{fig:clusters} top row).

Intuitively, we then select the final entity based on how often nearby points (those from clusters linked with an edge) were labeled with a sub- or superclass (ancestor-descendant semantic relationship).
Formally, we write $e\!\leq\!e'$ when entity $e$ is a subclass\footnote{By convention, an entity is a subclass of itself, \ie $e\leq e$} of $e'$, and $1[e\!\leq\!e']$ as the indicator function that takes value $1$ if, and only if, $e\!\leq\!e'$.
We then define the weight given to each entity $e\in v_i$ as:
{\small
\begin{equation}
  w(e, v_i)=\sum_{\substack{v_j\in G\\j\neq i}}\sum_{e'\in v_j}w_{i,j}\cdot\big(1[e\!\leq\!e']+1[e'\!\leq\!e]\big).
\end{equation}}%
Intuitively, $w(e, v_i)$ measures how likely it is for vertex $v_i$ to have meaning represented by $e$, based on how often nearby points were labeled with a sub- or superclass.

Finally, we choose as an entity for $v_i$ the one with the maximum weight among all entities contained in $v_i$.
Formally:
{\small
\begin{equation}
  e^*_{v_i}=\arg\max_{e\in v_i}w(e, v_i)
\end{equation}}%
if $\max_{e\in v_i}w(e, v_i)\neq0$ and \textit{undefined} otherwise (the cluster remains ambiguous).

\subsection{Post-processing unrecognized and ambiguous annotations}
\label{subsec:postproc}

The goal of the previous step is to assign a single entity to each vertex of the co-occurrence graph $G$ (Sec.~\ref{subsec:sense-disambiguation}): we refer to these vertices as \textit{unambiguous}.
However, there might be some vertices left that have multiple entities assigned to them, \ie{} the points contained in that vertex have different meanings.
We call these vertices \textit{ambiguous} (\eg{} $v_i$ in Fig.~\ref{fig:clusters}).
Finally, the unrecognized vertices do not have an assigned meaning (\eg{} $v_k$ in Fig.~\ref{fig:clusters}).

In the first post-processing step we assign an entity to the unrecognized vertices if there is an edge in $G$ connecting them to an unambiguous vertex.
We take the entity assigned to this unambiguous vertex as the meaning for the unrecognized ones.
With this assignment we are giving meaning to strings that were not in the language ontology, \eg{} the raw string \textit{spects} gets assigned the meaning \textit{Glasses} (Fig.~\ref{fig:clusters}).

Now we tackle the ambiguous vertices.
Often an ambiguous vertex has two meanings, one being a more general form of the other (\eg{} \textit{Cell phone} and \textit{iPhone}).
We remove the more specific one in these cases as it is the safer choice.
Other cases can be disambiguated by simple string matching between the raw string and the string representing the possible entities (\eg{}  the raw string \textit{house} gets matched to the entities \textit{House} and \textit{Dr. House}).
If we find an exact string match (\eg{} \textit{House}), we assign such entity to the vertex.

After post-processing, the number of ambiguous or unrecognized vertices becomes negligible in practice (Tab.~\ref{tab:disambiguation}), so it is feasible to annotate them manually or discard them.
We refer to the final set of unambiguous points as $\mathcal{U}$.

\subsection{Natural vocabulary emergence}
\label{subsec:natural_vocab}
The previous steps of our algorithm assign an ontology entity to each annotated point
(we discard the remaining ambiguous or unrecognized points).
We define the \textit{full natural vocabulary} $\mathcal{N}$ as the whole set of entities that have
been assigned to all points in our dataset.

Since the entities we use are part of a hierarchy of physical objects, all entities in this natural
vocabulary define a sub-hierarchy, which we refer to as \textit{natural visual hierarchy}.
Intuitively, it is a simplification of the physical object hierarchy
(Sec.~\ref{subsec:matching-to-a-language-vocabulary}),
excluding the entities that have never been annotated.
We argue that this makes the relationships between entities more natural and closer to the annotators knowledge,
\eg{} \textit{Dolphin} might become a direct subclass of \textit{Mammal}, instead of having the
classes \textit{Marine mammal} and \textit{Cetacea} in between.

Section~\ref{sec:learning-from-open-vocabulary-annotations} shows how we can leverage the natural
visual hierarchy directly with all its depth and breadth to learn a projection from visual features
to a textual embedding and perform zero-shot classification on $\mathcal{N}$.

\paragraph{Casting to a vocabulary of desired size:}
Having a smaller vocabulary can be advantageous in some applications.
In standard multi-class classification, for instance, having more samples per class and a smaller
vocabulary with more balanced classes are desirable properties (Fig.~\ref{fig:class_distribution}).
Given a desired vocabulary size $n < \left|\mathcal{N}\right|$, we propose to automatically find
the \textit{reduced natural vocabulary} $\mathcal{N}_n\subset\mathcal{N}$ that best represents the
full natural vocabulary.
On the one hand, we want this reduced vocabulary to cover as many of the classes from the full one
as possible, \ie{} as many classes as possible should be descendants of an entity in the reduced
vocabulary (\eg{} the class \textit{Animal} covers \textit{Mammal}, \textit{Dog} and
\textit{Chihuahua}).
On the other hand, we also want the reduced vocabulary to be as specific as possible, \ie{} we
want to reduce the number of hops in the hierarchy from all covered classes to the closest class
in the reduced vocabulary (\eg{} \textit{Dog} is preferred over \textit{Animal} to cover the class
\textit{Chihuahua}).

Let us formalize these two criteria.
For each entity $e\in\mathcal{N}$, we define $P(e)\subset\mathcal{U}$ as the set of unambiguous
points (Sec.~\ref{subsec:postproc}) whose meaning is associated to $e$.
Given a vocabulary $\mathcal{V}\subset\mathcal{N}$ we define the set of entities of
$\mathcal{N}$ covered by $\mathcal{V}$ as:
{\small
\begin{equation}
  \mathcal{C}(\mathcal{V})=\left\{e\in\mathcal{N}\ \big|\ \exists\, e'\in\mathcal{V} \text{ such that } e\leq e'\right\}.
\end{equation}}%
We then define \textit{coverage} as the percentage of points in $\mathcal{N}$ that are associated to
a sub-class of an entity in $\mathcal{V}$ (covered points):
{\small
\begin{equation}
  \mathrm{cover}(\mathcal{U},\mathcal{V})=\frac{\sum_{e\in\mathcal{C}(\mathcal{V})}|P(e)|}{|\mathcal{U}|},
\end{equation}}%
where $|\!\cdot\!|$ denotes set cardinality.
We then define \textit{specificity} as the percentage of covered points whose entity is actually in
the reduced vocabulary:
{\small
\begin{equation}
  \mathrm{spec}(\mathcal{U},\mathcal{V})=\frac{\sum_{e\in\mathcal{V}}|P(e)|}{\sum_{e\in\mathcal{C}(\mathcal{V})}|P(e)|}.
\end{equation}}%

Based on the above definitions, we can now define the optimal reduced natural vocabulary
$\mathcal{N}_n$ of size $n$ as:
{\small
\begin{equation}
  \mathcal{N}_n=\arg\max_{\substack{\mathcal{V}\subset\mathcal{N}\\|\mathcal{V}|=n}}\Big(\alpha\cdot\mathrm{cover}(\mathcal{U},\mathcal{V})+(1-\alpha)\cdot\mathrm{spec}(\mathcal{U},\mathcal{V})\Big),
\end{equation}}%
where $\alpha$ controls the trade-off between coverage and specificity.

We optimize this function using dynamic programming, taking advantage of the tree structure of the
vocabulary.
This produces a vocabulary of fixed size that is semantically as specific as possible while covering
as many classes as possible, with the option to adjust the weight between the two terms.

\subsection{Automatically making class labels more specific}
\label{subsec:specialization-method}

A consequence of our free-form protocol (Sec.~\ref{subsec:click-and-type}) is that annotators can use different degrees of specificity for naming objects, \eg{}  some might label all dogs simply as \textit{Dog}, while others might label specific breeds such as \textit{Labrador} or \textit{Chihuahua}.
We propose to make the generic annotations in a certain image $I$ more specific, by taking into account other raters with more knowledge in that field (although they have not labeled that particular image~$I$).
The key intuition is that more expert raters have labeled {\em other images} with more specific labels, which provides samples of the appearance of specific classes.
In our example, we can then use these samples to determine the breed of a dog annotated in $I$ as just {\em Dog}.

We employ a nearest neighbor approach based on the visual representation of a patch around an annotated point.
Formally, given a point $p$ annotated with entity $e_p\!\in\!\mathcal{N}$,
the aim is to predict which of its children
is the most probable.
We do so by looking for the point $p^*$
whose surrounding patch looks visually most similar to the patch around $p$, among the points $q$ labeled with a child class:
{\small
\begin{equation}
  p^* = \arg\!\!\!\!\!\!\min_{\substack{q\\e_q\text{ child of }e_p}} ||f(p) - f(q)||,
\end{equation}}%
where $f(\cdot)$ is a visual feature representation of the patch.
We then make the annotation of $p$ more specific as the entity $e^*_{p}$ associated with $p^*$.

We propose to use the ratio between the distance to the second nearest child class and the nearest child class as a confidence measure.
This allows us to decide whether we should make the label more specific, or leave it as it is.
We found this to work well in practice (Sec.~\ref{subsec:eval_specialization}).

  \section{Experimental Validation}
  \label{sec:experimental-validation}

  \subsection{Data collection and ontology choice}
  \label{subsec:data_collection}

  We annotate \att{\num{15100}} images from the \openimages{}~\cite{kuznetsova18arxiv} dataset, as they were randomly sampled without any prior vocabulary in mind.
  We had each image annotated by \att{\num{5}} different annotators from a pool of \att{\num{111}}, all of them based in India and fluent speakers of English.
  To study geographical biases,
  \att{\num{5100}} of these images were further annotated by \att{\num{5}} annotators from a pool of \att{\num{13}} living in the USA (American English).
  In total we collected \att{\num{729469}} point annotations.

  As the language ontology, we use an evolvement of FreeBase~\cite{bollacker08sigmod}.
  We use JFT~\cite{hinton14nips} as the hierarchy of physical objects,
  which is a curated visual hierarchy containing over \num{30000} classes referring to physical objects.
  It is very fine grained, covering very specific classes such as \textit{iPhone}, \textit{Greater Swiss mountain dog}, or \textit{Tulipa kolpakowskiana}.

  \subsection{From raw strings to ontology entities}
  \label{subsec:eval_normalization}

  \paragraph{Evaluation:}
  This section evaluates our approach for mapping the raw free-form annotation strings to entities in the language ontology (Sec.~\ref{subsec:matching-to-a-language-vocabulary}-\ref{subsec:sense-disambiguation}).
  We first analyze the percentage of annotated points whose meaning is unambiguous, ambiguous, or unrecognized (Sec.~\ref{subsec:postproc})
  at each of the steps (Table~\ref{tab:disambiguation}).
  We can see that the initial matching to the ontology (Sec.~\ref{subsec:matching-to-a-language-vocabulary}) already unambiguously recognizes \att{\num{84.4}\%} of the annotations.
  Click clustering (Sec.~\ref{subsec:matching-between-annotators}) and meaning disambiguation (Sec.~\ref{subsec:sense-disambiguation}) bring this percentage up to \att{\num{95.2}\%}.
  Finally, post-processing unrecognized points improves the disambiguated rate to \att{\num{96.5}\%} and the final post-processing of ambiguous points (Sec.~\ref{subsec:postproc}) makes it reach \att{\num{99.2}\%}.
  Hence, all proposed steps contribute to performance, and in the final outcome essentially all points are unambiguously assigned to one entity.

  \begin{figure}
    \centering
    \resizebox{\linewidth}{!}{%
    \begin{tikzpicture}[/pgfplots/width=1.45\linewidth, /pgfplots/height=0.9\linewidth]
  \begin{axis}[%
  ymax=1e5,
  ymin=0.8,
  xmax=1,
  ymode=log,
  xlabel=Ratio of classes,
  ylabel=Counts,
  xlabel shift={-2pt},
  ylabel shift={-3pt},
  font=\small,
  enlargelimits=false,
  clip=true,
  grid style=dotted, grid=both,
  major grid style={white!65!black},
  minor grid style={white!85!black},
  xtick={0,0.1,...,1.1},
  legend style={at={(0.01,0.01)},
  anchor=south west}]
    \putpin{(0.11392,13768)}{3mm}{45}{Woman}
    \putpin{(0.20253,10574)}{4mm}{45}{Shirt}
    \putpin{(0.30380,8816)}{4mm}{45}{Cloud}
    \putpin{(0.40506,5138)}{5mm}{45}{Trousers}
    \putpin{(0.50633,4284)}{1mm}{45}{Finger}
    \putpin{(0.59494,3098)}{6mm}{45}{Fence}
    \putpin{(0.70886,2853)}{2mm}{45}{Dress}
    \putpin{(0.81013,2393)}{6mm}{45}{Shrub}
    \putpin{(0.89873,2090)}{2mm}{45}{Stairs}
    \addplot+[blue,no markers, ultra thick] table[col sep = comma,x=rank,y=counts]
    {figures/data/class_histogram_80.csv};
    \addlegendentry{80 (Reduced)}

    \putpin{(0.10017, 2450)}{2mm}{45}{Mountain}
    \putpin{(0.20200, 1139)}{4mm}{45}{Earring}
    \putpin{(0.29549, 673)}{4mm}{45}{Button}
    \putpin{(0.40067, 414)}{4mm}{45}{Costume}
    \putpin{(0.50584, 245)}{2mm}{45}{Feather}
    \putpin{(0.59933, 172)}{1mm}{45}{Soldier}
    \putpin{(0.70451, 128)}{7mm}{45}{Christmas ornament}
    \putpin{(0.80467, 98)}{2mm}{45}{Twig}
    \putpin{(0.90150, 79)}{4mm}{45}{Beanie}

    \addplot+[red,no markers, ultra thick] table[col sep = comma,x=rank,y=counts]
    {figures/data/class_histogram_600.csv};
    \addlegendentry{600 (Reduced)}

    \putpin{(0.10000, 131)}{3mm}{45}{Thumb}
    \putpin{(0.20000, 37)}{3mm}{45}{Sausage}
    \putpin{(0.30000, 16)}{3mm}{45}{Berry}
    \putpin{(0.40000, 9)}{3mm}{45}{Extension cord}
    \putpin{(0.50000, 5)}{2mm}{45}{Goldfish}
    \putpin{(0.60000, 3)}{8mm}{45}{Brussels sprout}
    \putpin{(0.70000, 2)}{3mm}{45}{Tow truck}
    \putpin{(0.80000, 1)}{3mm}{60}{Zombie}
    \putpin{(0.90000, 1)}{7mm}{90}{Stuffed peppers}
    \addplot+[green,no markers, ultra thick, mark=none] table[col sep = comma,x=rank,y=counts]
    {figures/data/class_histogram_full_4020.csv};
    \addlegendentry{4020 (Full)}
    \label{fig:class_distribution:full}
  \end{axis}
\end{tikzpicture}
    }
    \caption{\textbf{Frequency of class labels}: Comparison of our different vocabularies (Full and reduced at 80 and 600 classes).
    \vspace{-6pt}}
    \label{fig:class_distribution}
  \end{figure}
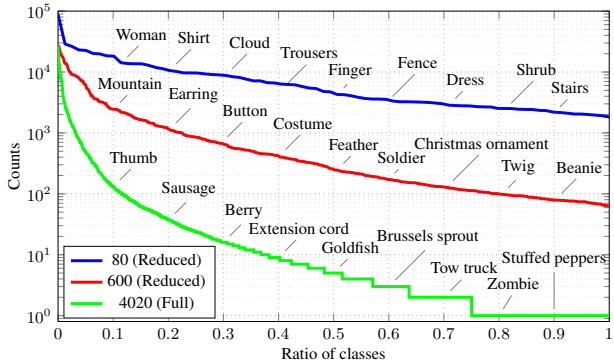
  \begin{table}[t]
    \resizebox{\linewidth}{!}{%
    \begin{tabular}{lrrrr}
      \toprule
      & Matching to & Click clust. +& Post-proc. &Post-proc.\\
      & ontology & Disambiguation & unrecognized&ambiguous\\
      \midrule
      Unambig. points (\%) & 84.4 & 95.2 & 96.5 & 99.2\\
      Ambig. points (\%) & 13.8 & 3.0 & 3.1 & 0.4\\
      Unrecog. points (\%) & 1.8 & 1.8 & 0.4 & 0.4\\
      \midrule
      Unambig. points & \num{615751} & \num{694343} & \num{703534} & \num{723545}\\
      Unambig. classes & \num{3772} & \num{3809} & \num{3809} & \num{4020}\\
      \bottomrule
    \end{tabular}}\\
    \caption{\textbf{Evaluation of the matching to an ontology}: Percentage of unambiguous, ambiguous, and unrecognized annotations (top three rows) and number of unambiguous points and classes (bottom two rows).
   }
    \label{tab:disambiguation}
  \end{table}

  Table~\ref{tab:disambiguation} also shows the number of point annotations assigned to a single
  class (entity)  and the total number of distinct such classes over all points.
  The final number of unambiguous classes is \att{\num{4020}}.

  We also evaluate the accuracy of our method for mapping raw input strings to entities in the language ontology.
  For this, we take a random sample of \att{\num{100}} raw strings that our method matches unambiguously to a single entity and manually verify whether they are correct.
  We obtain \att{\num{89}} correct matches, the remaining being:
  (a) \att{\num{5}} polysemic words that are present in our dataset with at least two of the meanings (\eg{} {\em mouse} meaning either \textit{Computer mouse} or \textit{Animal mouse}),
  (b) \att{\num{5}} wrong matches for which there is a correct entity in JFT (\eg{} {\em shades} matched to \textit{Shade}, not \textit{Sunglasses}), and \att{\num{1}} that has no corresponding entity in JFT  (\textit{monster}).

  For reference,~\cite{russell08ijcv} manually matched the free-form labels of the LabelMe dataset to the WordNet ontology and found a matching entity for 93\% of all classes.
  Thus, we can conclude that the used hierarchy of physical objects~\cite{hinton14nips} is very complete and that our approach for matching into the ontology is accurate.

  \paragraph{Exhaustiveness:}
  Each annotator provides labels from an average of \att{\num{7.53}} different classes per image, well above the number of classes per image in \openimages{} (\att{\num{2.9}} out of \att{\num{20000}} classes) and \coco{} (\att{\num{3.5}} out of 80 classes).
  Since various annotators label the same images in our experiments, we can compute the final classes as the union of classes on which at least a certain percentage of the annotators agree.
  Table~\ref{tab:labels_per_image} shows the mean number of distinct classes obtained per image.
  At \att{\num{50}\%} agreement, we obtain \att{\num{5.4}} different classes per image, up to \att{\num{19.7}} when we consider the set union among all annotators.

  \begin{table}[t]
    \centering
    \resizebox{\linewidth}{!}{%
    \begin{tabular}{l@{\hspace{25mm}}rrrrrrrrrr}
      \toprule
      Annotator agreement & \textit{Any} & 25\% & 50\% & 75\%  & \textit{All} \\
      \midrule
      Labels per image & 19.7 & 9.0 & 5.4 & 3.0 & 1.4 \\
      \bottomrule
    \end{tabular}}\\[2mm]
    \caption{\textbf{Mean number of classes per image} at different levels of per-image annotator agreement for a class to be considered annotated;
    from the set union of the class labels from any annotator (\textit{Any}), to full agreement between all annotators (\textit{All}).}
    \label{tab:labels_per_image}
  \end{table}

  \paragraph{Class distribution:}
  We analyze the number of examples per class in Fig.~\ref{fig:class_distribution} (\ref{fig:class_distribution:full}) and find that it follows a long-tail distribution, where \att{\num{74.2}\%} of points are covered by the \att{\num{100}} most frequent classes.
  The 5 most frequent classes are
  \textit{Tree},
  \textit{Sky},
  \textit{Man},
  \textit{Hair}, and
  \textit{Wall}.
  We observe that both \textit{thing} and \textit{stuff} classes~\cite{heitz08eccv,caesar18cvpr} get naturally labeled.

  Apart from these common classes, annotators also provide very specific classes such as
  \textit{Dreadlocks},
  \textit{Hijab},
  \textit{Astronaut},
  \textit{Submarine}, or
  \textit{Batman}.
  This highlights the benefit of our approach: annotators naturally and effortlessly label
  objects that are rare and, if they know their specific name, provide it.
  Our annotations thus show high specificity and diversity and cover classes that
  were previously missing from standard recognition datasets.

  \paragraph{Knowledge discovery:}
  Our method can match unrecognized words to unambiguous annotations, which can be used to enrich the language ontology (Sec.~\ref{subsec:postproc}).
  The most frequent example is \textit{spects} (India) and \textit{specs} (USA), which are both matched to \textit{Glasses} but interestingly have different dominant spellings depending on the geographical origin.
  Other discovered words from the annotators in India include \textit{jerkin} as \textit{Jacket} and \textit{glouse} as \textit{Glove};
  and from the annotators in the USA \textit{slacks} as \textit{Trousers} and \textit{steps} as \textit{Stairs}.

\begin{figure}[t]
    \centering
    \resizebox{\linewidth}{!}{%
    \includegraphics{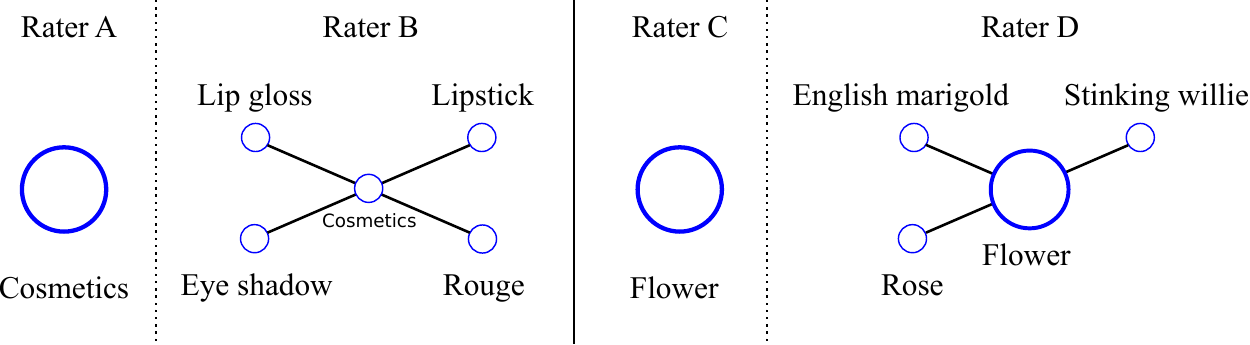}
    }\\[2mm]
    \caption{\textbf{Examples of annotator specific knowledge}: Central bubbles represent generic entities, and the satellite ones more specific classes.
    Size encodes frequency.
    We find that some annotators are more specific than others in the
    object names they provide for different classes of objects.
    \vspace{-8pt}}
    \label{fig:specificity_examples}
  \end{figure}

  Our method also discovers the biases and knowledge of the annotators.
  We see, for example, that some annotators know specific flower types and label \textit{English marigold} or \textit{Stinking willie} while others use the more generic entity \textit{Flower} (Fig.~\ref{fig:specificity_examples}).
  Section~\ref{subsec:eval_specialization} evaluates how we use this information to make the annotations more specific.

  \subsection{Reduced natural vocabulary}
  \label{subsec:eval_natural_vocab}
\begin{figure}
      \centering
      \resizebox{0.95\linewidth}{!}{%
      \begin{tikzpicture}[/pgfplots/width=1.25\linewidth,/pgfplots/height=0.75\linewidth]
  \begin{axis}[%
  ymin=25,ymax=100,xmin=25,xmax=100,
  ylabel=Coverage,
  xlabel=Specificity,
  xlabel shift={-2pt},
  ylabel shift={-3pt},
  font=\small,
  axis equal image=false,
  enlargelimits=false,
  clip=true,
  axis on top=false,
  grid style=dotted, grid=both,
  major grid style={white!65!black},
  minor grid style={white!85!black},
  xtick={0,10,...,100},
  ytick={0,10,...,100},
  minor xtick={0,2,...,100},
  minor ytick={0,2,...,100},
  legend style={at={(0.99,0.01)},
  anchor=south east}]

    \addplot+[red,solid,ultra thick,mark=none] table[col sep = comma,x=specificity,y=coverage] {figures/data/5k+100_aus+10k+5k+100_hyd_spec_cover_600.csv};
    \addlegendentry{Natural vocabulary (600)}
    \addplot+[red,only marks,mark=*,mark options={solid,fill=red}] table[col sep = comma,x=specificity,y=coverage] {figures/data/5k+100_aus+10k+5k+100_hyd_spec_cover_oid.csv};
    \addlegendentry{\openimages{} (600)}
    \label{fig:coverage_specificity:oid}
    \addplot+[blue,solid,ultra thick,mark=none] table[col sep = comma,x=specificity,y=coverage] {figures/data/5k+100_aus+10k+5k+100_hyd_spec_cover_80.csv};
    \addlegendentry{Natural vocabulary (80)}
    \addplot+[blue,only marks,mark=square*,mark size=1.7,mark options={solid,fill=blue}] table[col sep = comma,x=specificity,y=coverage] {figures/data/5k+100_aus+10k+5k+100_hyd_spec_cover_coco.csv};
    \addlegendentry{\coco{} (80)}
    \label{fig:coverage_specificity:coco}
    \addplot+[black,only marks,mark=triangle*,mark size=2.2,mark options={solid,fill=black}] table[col sep = comma,x=specificity,y=coverage] {figures/data/5k+100_aus+10k+5k+100_hyd_spec_cover_coco_stuff.csv};
    \addlegendentry{\cocostuff{} (147)}
    \label{fig:coverage_specificity:cocostuff}
  \end{axis}
\end{tikzpicture}
      }
      \caption{\textbf{Natural \vs predefined vocabulary}: Comparison in terms of coverage and specificity.
      The reduced natural vocabulary works well in all ranges of coverage and specificity, and reaches significantly better values than the predefined vocabularies of \coco{} and \openimages{}.
      \vspace{-10pt}}
      \label{fig:coverage_specificity}
    \end{figure}
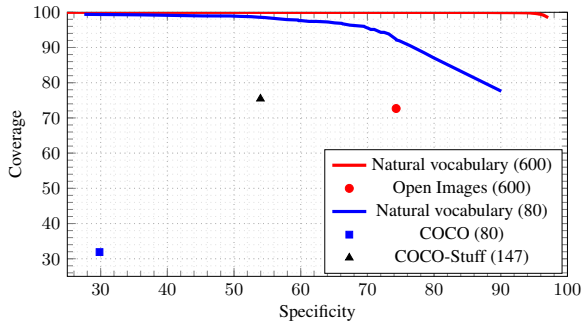

  This section quantitatively evaluates the reduced natural vocabulary $\mathcal{N}_n$ that maximizes coverage and specificity (Sec.~\ref{subsec:natural_vocab}).
  We sweep the values of the trade-off parameter $\alpha$ for $n\!=\!80$ and $n\!=\!600$, obtaining a curve in the coverage-specificity plane (Fig.~\ref{fig:coverage_specificity}).
  We compare to three predefined vocabularies: 80 \coco{} classes (\ref{fig:coverage_specificity:coco}), the 600 classes of \openimages{} for which there are box annotations~\cite{kuznetsova18arxiv}~(\ref{fig:coverage_specificity:oid}), and the \att{147} classes from \cocostuff{}~\cite{caesar18cvpr} that are represented in JFT\footnote{This leaves out \att{22} classes that are either too broad (\eg{} \textit{solid-other}) or are a combination of class and attribute (\eg{} \textit{floor-stone}).}~(\ref{fig:coverage_specificity:cocostuff}).
  The three of them have significantly less coverage and specificity than our natural vocabulary.
  Surprisingly, even a learnt natural vocabulary of only 80 classes has higher coverage and
  specificity than the \openimages{} vocabulary of 600 classes, despite the latter was designed
  to cover important classes in the \openimages{} dataset~\cite{kuznetsova18arxiv}.
  This highlights the importance of learning a natural vocabulary, rather than
  using a predefined vocabulary based on an educated guess of what classes are important and frequent.
  The comparison to \cocostuff{} shows that the better coverage and specificity does not come only from having both \textit{things} and \textit{stuff} in the vocabulary.

  While our \textit{full natural vocabulary} covers a large number of classes (\att{\num{4020}}), many of them have few examples (Sec.~\ref{subsec:eval_normalization}).
  Standard multi-class classification~\cite{krizhevsky12nips} however assumes somewhat balanced class distributions, with a significant number of examples per class.
  In Fig.~\ref{fig:class_distribution} we show the class frequency of the full and reduced natural vocabularies.
  We find that by constructing a reduced natural vocabulary (Sec.~\ref{subsec:natural_vocab}), it becomes more balanced, despite not explicitly optimizing for that.
  In the full vocabulary, \att{\num{62}\%} of all classes have $<$\att{\num{10}} points associated.
  On the reduced natural vocabulary of 80 classes, all classes have $\geq$\att{\num{100}} examples.

  \subsection{Making class labels more specific}
  \label{subsec:eval_specialization}

  Different annotators provide labels at different levels of specificity, reflecting their knowledge of particular domains (Fig.~\ref{fig:specificity_examples}).
  We evaluate here the performance of our method in predicting specific labels for points annotated with more generic ones (Sec.~\ref{subsec:specialization-method}).

  \paragraph{Data:}
  In order to evaluate the accuracy of predicting more specific labels, we create an evaluation set from our data.
  We select all classes which were labeled at least \att{\num{20}} times and have at least 2 children classes with at least \att{\num{2}} examples each.
  With this procedure, we obtain a dataset consisting of \att{\num{398000}}
  annotations.
  For each each point $p$ with its annotated class $e_p$, we pretend that $p$ is annotated with its parent's label and evaluate the ability of our procedure to correctly predict $e_p$ as the most likely child class.
  To obtain the feature representation $f(p)$, we extract a patch of $120\!\times\!120$ pixels around $p$ and represent it using the activations of the last layer of a ResNet~\cite{he16cvpr} model pre-trained on ImageNet~\cite{russakovsky15ijcv}.

  \paragraph{Results:}
  Our method predicts more specific label correctly for \att{83.6\%} of the examples.
  As a baseline, always selecting the most frequent child class leads to only \att{58.5\%} accuracy.
  This demonstrates that visual cues help finding more specific classes with high accuracy.

\begin{figure}[t]
    \centering
    \resizebox{1\linewidth}{!}{%
    \begin{tikzpicture}[/pgfplots/width=1.35\linewidth, /pgfplots/height=1.35\linewidth]
  \begin{axis}[%
  ymin=0.5,ymax=1,xmin=0,xmax=1,
  xlabel=Percentage of points for which we made the label more specific,
  ylabel=Accuracy,
  xlabel shift={-2pt},
  ylabel shift={-3pt},
  font=\small,
  axis equal image=true,
  enlargelimits=false,
  clip=true,
  grid style=dotted, grid=both,
  major grid style={white!65!black},
  minor grid style={white!85!black},
  xtick={0,0.1,...,1.1},
  ytick={0,0.1,...,1.1},
  minor xtick={0,0.02,...,1},
  minor ytick={0,0.02,...,1},
  legend style={at={(0.05,0.05)},
  anchor=south west}]
    \addplot+[red,no markers, ultra thick] table[col sep = comma,x=ratio_specialized,y=accuracy] {figures/data/specialization_k-nn.csv};
    \addlegendentry{Ours}
    \addplot+[blue,no markers, ultra thick] table[col sep = comma,x=ratio_specialized,y=accuracy] {figures/data/specialization_most_frequent_child.csv};
    \addlegendentry{Most frequent child}

  \end{axis}
\end{tikzpicture}
    }
    \caption{\textbf{Making point labels more specific}: Comparison of our approach and the baseline.
    \vspace{-8pt}}
    \label{fig:specialization_results}
  \end{figure}
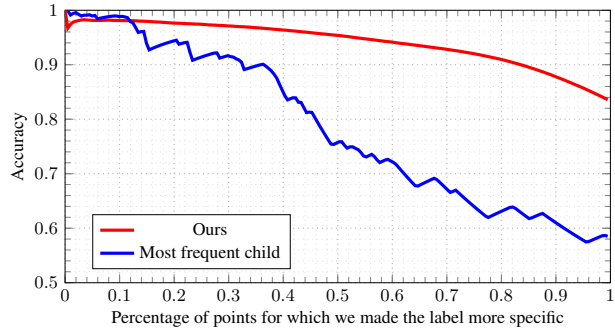

  Since our method also produces a confidence value, we can use that to only make classes more specific if the model is confident.
  Fig.~\ref{fig:specialization_results} shows the accuracy of our approach when making the label of a varying percentage of points more specific.
  As can be seen, our method produces a meaningful confidence measure, allowing to make the label of \att{50\%} of all points more specific with \att{95\%} accuracy.

  \subsection{Annotating a predefined vocabulary}
  \label{subsec:eval_closed_vocab}
  To further highlight the usefulness of our annotation strategy, we evaluate it as a way of labeling
  the classes of a predefined vocabulary.
  Specifically, we take the subset of 100 images from \openimages{} that were manually densely
  annotated by experts with image-level labels for the \num{600} so-called \textit{boxable}
  classes~\cite{kuznetsova18arxiv}.
  We then obtain image-level annotations using click \& type (we disregard the location of the points)
  at different level of agreement between the \num{5} annotators available and ignoring
  annotations out of the predefined \num{600} classes

\begin{figure}
    \centering
    \resizebox{\linewidth}{!}{%
    \begin{tikzpicture}[/pgfplots/width=1.4\linewidth, /pgfplots/height=1.4\linewidth]
  \begin{axis}[%
  ymin=0.8,ymax=1,xmin=0.1,xmax=0.7,
  xlabel=Recall,
  ylabel=Precision,
  xlabel shift={-2pt},
  ylabel shift={-3pt},
  font=\small,
  axis equal image=true,
  enlargelimits=false,
  clip=true,
  grid style=dotted, grid=both,
  major grid style={white!65!black},
  minor grid style={white!85!black},
  xtick={0,0.1,...,1.1},
  ytick={0,0.1,...,1.1},
  minor xtick={0,0.01,...,1},
  minor ytick={0,0.01,...,1},
  legend style={at={(0.05,0.05)},
  anchor=south west}]

    \foreach \f in {0.1,0.2,...,0.9}{%
    \addplot[white!50!green,line width=0.2pt,domain=(\f/(2-\f)):1,samples=200,forget plot]{(\f*x)/(2*x-\f)};
    }

    \addplot+[red,no marks, ultra thick, forget plot] table[x=recall,y=precision] {figures/data/oid_coverage_agreement.txt};
    \label{plot:cover:natural}

    \addplot+[red,only marks,mark=*, mark size=1.9, thick, mark options={solid,fill=red}, forget plot] table[x=recall,y=precision] {figures/data/oid_coverage_mean.txt};
    \label{plot:cover:mean}

    \addlegendimage{no markers,red, ultra thick}
    \addlegendentry{Natural vocabulary - Agreement between annotators}
    \addlegendimage{red,only marks,mark=*}
    \addlegendentry{Natural vocabulary - Mean of annotators}
  \end{axis}
\end{tikzpicture}
    }
    \caption{\textbf{Precision-recall on 100 images of \openimages{}}: Our annotation strategy recovers up to \att{\num{67}\%} of the dense annotations.
    \vspace{-8pt}}
    \label{fig:closed_vocab_cover}
  \end{figure}
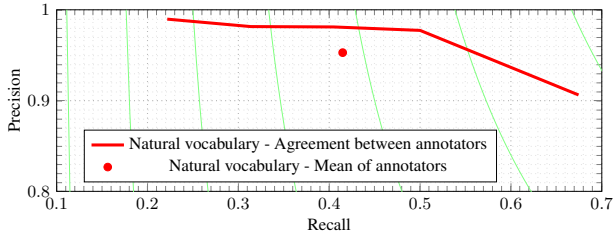

  We evaluate the image-level annotations against the dense annotations by the experts,
  which we consider as ground truth.
  Figure~\ref{fig:closed_vocab_cover} shows the precision-recall plot at different levels of
  agreement between annotators (from 1 to~5).
  Using the agreement between annotators leads to consistently better results than the mean
  performance over the 5 annotators (\ref{plot:cover:mean}).
  The mean precision of the annotated concepts is \att{\num{95}\%}, considerably better than the 83\% reported for COCO~\cite{lin14eccv}.
  This highlights the advantage of free-form annotation: annotators are not forced to assign pre-defined class names to objects which might bias the labeling (\eg{} if an alpaca is shown to an annotator but only \textit{Sheep} is in the vocabulary, then they might have a tendency to wrongly label it as such).
  Instead, they can name objects with concepts they are confident to use, hence making them more precise (Sec.~\ref{subsec:click-and-type}).

  While we cover the predefined vocabulary of \openimages{} well, our annotations are not limited to it.
  In particular, out of the \att{\num{516}} classes our procedure annotated in these images, \att{\num{346}} can be matched to in-vocabulary \openimages{} classes, while \att{\num{170}} (\att{\num{33}\%}) are out of the predefined vocabulary.
  The most common examples of such out-of-vocabulary classes are \textit{Sky}, \textit{Sign}, \textit{Wall}, and \textit{Light fixture}.

  Within the predefined vocabulary, our annotations include more specific classes than those in \openimages{}.
  As an example, in these 100 images our annotators labeled \textit{Capo}, \textit{Cymbal}, and \textit{Tuba}, all of them more specific classes under \textit{Musical instrument} (and absent in the 600 boxable classes of \openimages{}).

  \section{Learning from Natural Vocabulary}\label{sec:learning-from-open-vocabulary-annotations}

  Annotators naturally provide labels that are more diverse and more specific compared to
  a large but pre-defined vocabulary such as the one of \openimages{}~\cite{kuznetsova18arxiv} (Sec.~\ref{subsec:eval_closed_vocab}).
  We conjecture that this increased diversity will allow to learn better embeddings. %
  To test this hypothesis we train a textual-visual embedding model~\cite{frome13nips,socher13nips,karpathy15cvpr} on our data.
  Thereby we consider all examples that have class labels which are either \openimages{} classes or children of them.
  When training the embedding we represent the class labels either by
  (i) using the class name directly or (ii) mapping the class name to its parent in the \openimages{} vocabulary and using that class instead.
  Finally, we use these two variants for zero-shot classification, where a better embedding is expected to generalize better.
  Next, we will describe the model and experimental setup in detail, before presenting results.

\paragraph{Visual-semantic embedding model:}
\label{subsec:vis-sem-model}
  We adopt the DeViSE model~\cite{frome13nips}, which learns a projection of textual and visual input into the same vector space.
  The model relies on a semantic embedding of words which is pre-trained in an unsupervised way~\cite{mikolov13iclr}.
  During training, the model is trained to predict a similar vector representation for an image and its class label,~\ie the image representation is aligned with the semantic text embedding.
  At test time, class labels are predicted based on the similarity between an image embedding and the embeddings of the class names.
  Such a model can be used for zero-shot classification by simply changing the class vocabulary at test time.

  During training, we learn a projection matrix $\mathbf{M}$, which maps image features %
  into the vector space of the textual embedding.
  We follow~\cite{frome13nips} and minimize a hinge rank loss function %
  using stochastic gradient descent.
  This loss encourages the similarity between matching (image, label) pairs %
  to be larger than that of the image and a negative label, by some margin. %
  As a negative label, we randomly sample a class from the \openimages{} vocabulary which violates the margin and thus leads to a non-zero loss.

\paragraph{Experimental setup:}\label{subsec:devise-results}
  To train and evaluate our method we use the 15100 images annotated as described in Sec.~\ref{subsec:data_collection}.
  We use \att{\num{12100}} images for training, and \att{\num{1500}} for test and validation, each.
  On the training set there are \att{\num{390000}} (point, label) pairs.
  For each of those examples we extract a patch of \att{$120\!\times\!120$} pixels around the click position to compute the image representation. %
  We use the last layer of a ResNet~\cite{he16cvpr} model pre-trained on ImageNet~\cite{russakovsky15ijcv} as feature extractor. %
  For obtaining text embeddings of class labels we use the word2vec model~\cite{mikolov13iclr} pre-trained on the Google News dataset.

  At test time, we use use the embedding model for zero-shot classification, where the objective is to correctly classify image patches into previously unseen classes.
  To create this set of unseen classes, we sample \att{300} classes from the test set (both \openimages{} and more specific classes). Then, we remove examples with these classes from the training set, so that they
  are not seen during training.

  \paragraph{Performance metrics:}
  We report hit@$k$,~\ie{} the percentage of cases for which the model returns the true label in its $k$ highest scoring predictions, averaged over all test classes.

\paragraph{Results:}
  Training with specific labels leads to significant performance gains in zero-shot classification (Tab.~\ref{tab:devise_results}).
  We see particularly high gains when testing on a vocabulary that combines zero-shot and the \openimages{} classes.
  We conclude that the increased specificity and diversity of our data allows to learn better embeddings compared to using
  a pre-defined and thus smaller vocabulary.
  \begin{table}[t]
    \resizebox{\linewidth}{!}{%
    \begin{tabular}{Hl@{\hspace{30pt}}l@{\hspace{40pt}}H@{\hspace{10pt}}r@{\hspace{10pt}}r@{\hspace{10pt}}r@{\hspace{10pt}}rH}
  \toprule
  & Vocabulary & \# Classes & Accuracy &    $k=$\ \ \ \ \ 1 & 2 & 5 & 10 & 20 \\
  \midrule
  & & & & \multicolumn{5}{c}{\openimages{} classes} \\
  \midrule
  specific\_VOC:unseen+OI\_zero\_shot\_per\_class & Zero-shot + OI & 779 & class & 0.0 & 0.6 & 3.5 & 7.9 & 17.9 \\
  specific\_VOC:unseen\_zero\_shot\_per\_class    &     Zero-shot &       300 &    class &  6.0 & 11.0 & 19.8 & 28.6 & 38.5 \\

  \midrule
  & & & & \multicolumn{5}{c}{Specific classes} \\
  \midrule
  specific\_VOC:unseen+OI\_zero\_shot\_per\_class & Zero-shot + OI & 779 & class &  \textbf{1.8} &  \textbf{4.4} & \textbf{11.7} & \textbf{19.9} & 29.2 \\
  specific\_VOC:unseen\_zero\_shot\_per\_class    &     Zero-shot &       300 &    class &  \textbf{7.7} & \textbf{13.7} & \textbf{22.4} & \textbf{31.2} & \textbf{42.4} \\
  \bottomrule
\end{tabular}\\
    }\\
    \caption{\textbf{Zero-shot classification results.}
    We observe that training with specific class labels outperforms training with classes from the \openimages{} vocabulary.
    \vspace{-8pt}}
    \label{tab:devise_results}
  \end{table}

  \section{Conclusion}
  We proposed an approach for annotating object classes using free-form text annotations.
  We first transform the raw strings into entities in an ontology of physical objects and then use co-occurrences in the images to disambiguate meanings. 
  Our method makes a \textit{natural vocabulary} emerge, which covers a large number of classes and makes the knowledge and biases of the annotators arise.
  We also automatically extract natural vocabularies of reduced size that have high object coverage while remaining specific.
  Free-form labeling is natural for annotators, they intuitively provide very specific and exhaustive annotations, and no training stage is necessary.
  Free-form labeling also builds a powerful synergy with speech-based interfaces~\cite{gygli19cvpr}, liberating the former from slow keyboard inputs, and the latter from requiring training annotators to use a predefined vocabulary.
  We believe that our method can lead to a paradigm shift in the way object classes are annotated.

   \section*{Acknowledgments}
   The icons in Fig.~\ref{fig:overview} are based on an original design by macrovector, Freepik.

{\small
\bibliographystyle{ieee}
\bibliography{shortstrings,loco}
}

\end{document}